\documentclass{article}

\usepackage{arxiv}

\usepackage[utf8]{inputenc} 
\usepackage[T1]{fontenc}    
\usepackage{hyperref}       
\usepackage{url}            
\usepackage{booktabs}       
\usepackage{amsfonts}       
\usepackage{nicefrac}       
\usepackage{microtype}      
\usepackage{lipsum}
\usepackage{soul}
\usepackage{tcolorbox}
\usepackage{times}
\usepackage{latexsym}
\usepackage{times}  
\usepackage{graphicx}  
\usepackage{epsfig}
\usepackage{multirow}
\usepackage{booktabs}
\usepackage{amssymb,amsmath}
\usepackage{epsfig}
\usepackage{algorithm,algorithmic}
\usepackage{xcolor,colortbl}
\usepackage{booktabs}
\usepackage{xcolor,colortbl}
\usepackage{makecell}
\usepackage{subcaption}
\usepackage{lipsum}
\usepackage{balance}
\title{An Editorial Network for Enhanced Document Summarization}

\author{
  Edward Moroshko\thanks{Work was done during a summer internship in IBM Research.} \\
  Electrical Engineering Dept.\\
  Technion -- Israel Institute of Technology\\
  Haifa, Israel \\
  \texttt{edward.moroshko@gmail.com} \\
   \And
  Guy Feigenblat, Haggai Roitman, David Konopnicki \\
  IBM Research\\
  Haifa University Campus\\
  Haifa, Israel \\
  \texttt{\{guyf,haggai,davidko\}@il.ibm.com} \\
  }

\date{}

\begin{document}
\maketitle
\begin{abstract}
We suggest a new idea of \textit{Editorial Network} -- a mixed extractive-abstractive summarization approach, which is applied as a post-processing step over a given sequence of extracted sentences. Our network tries to imitate the decision process of a human editor during summarization. Within such a process, each extracted sentence may be either kept untouched, rephrased or completely rejected. We further suggest an effective way for training the ``editor" based on a novel soft-labeling approach. Using the CNN/DailyMail dataset we demonstrate the effectiveness of our approach compared to state-of-the-art extractive-only or abstractive-only baseline methods.
\end{abstract}

\section{Introduction}\label{sec:intro}
Automatic text summarizers condense a given piece of text into a shorter version (the summary). This is done while trying to preserve the main essence of the original text and keeping the generated summary as readable as possible.

Existing summarization methods can be classified into two main types, either \textit{extractive} or \textit{abstractive}~\cite{Gambhir:2017}. Extractive methods select and order text fragments (e.g., sentences) from the original text source~\cite{Cheng2016ACL,Dlikman2016UsingML,DBLP:conf/emnlp/DongSCHC18,summit2017,Nallapati2017a,ZhangLatent2018}. Such methods are relatively simpler to develop and keep the extracted fragments untouched, allowing to preserve important parts, e.g., keyphrases, facts, opinions, etc. Yet, extractive summaries tend to be less fluent, coherent and readable and may include superfluous text.

Abstractive methods apply natural language paraphrasing and/or compression on a given text. A common approach is based on the encoder-decoder (seq-to-seq) paradigm~\cite{Sutskever2014Seq}, with the original text sequence being encoded while the summary is the decoded sequence. While such methods usually generate summaries with better readability, their quality declines over longer textual inputs, which may lead to higher redundancy~\cite{Paulus2017ADR}. Moreover, such methods are sensitive to vocabulary size, making them more difficult to train and generalize~\cite{See2017GetTT}.

A common approach for handling long text sequences in abstractive settings is through \textit{attention} mechanisms, which aim to imitate the attentive reading behaviour of humans~\cite{Chopra2016Abs}.
Two main types of attention methods may be utilized, either \textit{soft} or \textit{hard}. Soft attention methods first locate salient text regions within the input text and then bias the abstraction process to prefer such regions during decoding~\cite{Cohan2018ADA,Gehrmann2018BottomUpAS,HsuAUM2018,DBLP:conf/conll/NallapatiZSGX16,Li2018KIGN,Pasunuru2018a,Tan2017a}. On the other hand, hard attention methods perform abstraction only on text regions that were initially selected by some extraction process~\cite{ChenFastAS2018,Nallapati2017a,liu2018a}.

Compared to previous works, whose final summary is either entirely extracted or generated using an abstractive process, in this work, we suggest a new idea of  ``\textit{Editorial Network}" (EditNet) -- a \textit{mixed extractive-abstractive} summarization approach.  A summary generated by \textit{EditNet} may include sentences that were either extracted, abstracted or of both types. Moreover, per considered sentence, \textit{EditNet} may decide not to take either of these decisions and completely reject the sentence. 

Using the CNN/DailyMail dataset we demonstrate that, \textit{EditNet}'s summarization quality transcends that of state-of-the-art abstractive-only baselines. \textit{EditNet}'s summarization quality is also demonstrated to be highly competitive with that of \textit{NeuSum}~\cite{Qingyu2018}, which is, to the best of our knowledge, the best performing extractive-only baseline. Yet, while \textit{EditNet} obtains (more or less) a similar summarization quality to that of \textit{NeuSum}, compared to the latter which applies only extraction, the former (on average) applies abstraction to the majority of each summary's extracted sentences.

\section{Editorial Network}
\begin{figure*}[t!]
  \centering
  \includegraphics[width=4.00in]{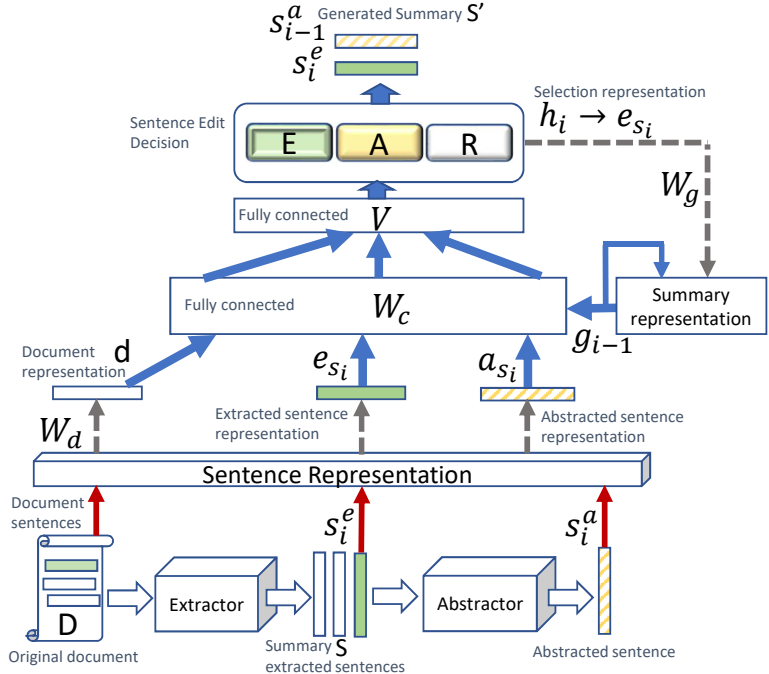}
  \caption{Editorial Network}\label{fig:architecture}
\end{figure*}

Figure~\ref{fig:architecture} now depicts the architecture of our proposed \textit{Editorial Network}-based approach.
We apply this approach as a post-processing step over a given summary whose sentences were selected by some extractor. The key idea is to try to imitate the decision process of a human editor who needs to edit the summary so as to enhance its quality.

Let $S$ denote a summary which was extracted from a given text (document) $D$.
The editorial process is implemented by iterating over sentences in $S$ according to the selection order of the extractor.
For each sentence in $S$, the ``editor" may make three possible decisions. The first decision is to keep the extracted sentence untouched (represented by label \textsf{E} in Figure~\ref{fig:architecture}). 
The second alternative is to rephrase the sentence (represented by label \textsf{A} in Figure~\ref{fig:architecture}). Such a decision, for example, may represent the editor's wish to simplify or compress the original source sentence. The last possible decision is to completely reject the sentence (represented by label \textsf{R} in Figure~\ref{fig:architecture}). For example, the editor may wish to ignore a superfluous or duplicate information expressed in the current sentence.
An example mixed summary generated by our approach is depicted in Figure~\ref{example summary}, further emphasizing the various editor's decisions.

\begin{figure}[tbh]
\begin{tcolorbox}[colback = white,boxrule=0.5pt]
  \textbf{Editor's automatic summary}:\\
  \textsf{E}: what was supposed to be a fantasy sports car ride at walt disney world speedway turned deadly when a lamborghini crashed into a guardrail.
  \textsf{A}: \textit{the crash took place sunday at the exotic driving experience}\footnote{Original extracted sentence: ``the crash took place sunday at the exotic driving experience , which bills itself as a chance to drive your dream car on a racetrack".}.
  \textsf{A}: \textit{the lamborghini 's passenger , gary terry , died at the scene}\footnote{Original extracted sentence: ``the lamborghini 's passenger , 36-year-old gary terry of davenport , florida , died at the scene , florida highway patrol said"}.
  \textsf{R}: \st{petty holdings , which operates the exotic driving experience at walt disney world speedway , released a statement sunday night about the crash.}
\end{tcolorbox}
\begin{tcolorbox}[colback = white,boxrule=0.5pt]
  \textbf{Ground truth summary}:\\
  the crash occurred at the exotic driving experience at walt disney world speedway. officials say the driver , 24-year-old tavon watson , lost control of a lamborghini. passenger gary terry , 36 , died at the scene.
\end{tcolorbox}\caption{An example mixed summary (annotated with the editor's decisions) taken from the CNN/DM dataset}\label{example summary}
\end{figure}

\subsection{Implementing the editor's decisions}
For a given sentence $s\in{D}$, we now denote by $s^{e}$ and $s^{a}$ its original (extracted) and paraphrased (abstracted) versions.
To obtain $s^{a}$ we use an abstractor, whose details will be shortly explained (see Section~\ref{sec:EA sum}). Let $e_{s}\in\mathbb{R}^{n}$ and $a_{s}\in\mathbb{R}^{n}$ further denote the corresponding sentence representations of $s^{e}$ and $s^{a}$, respectively. Such representations allow to compare both sentence versions on the same grounds.

Recall that, for each sentence $s_{i}\in{S}$ (in order) the editor makes one of the three possible decisions: extract, abstract or reject $s_i$.
Therefore, the editor may modify summary $S$ by paraphrasing or rejecting some of its sentences, resulting in a mixed extractive-abstractive summary $S'$.

Let $l$ be the number of sentences in $S$. In each step $i\in\{1,2,\ldots,l\}$, in order to make an educated decision, the editor considers both sentence representations $e_{s_i}$ and $a_{s_i}$ as its input, together with two additional auxiliary representations. The first auxiliary representation is that of the whole document $D$ itself, hereinafter denoted $d\in\mathbb{R}^{n}$. Such a representation provides a \textit{global context} for decision making. Assuming document $D$ has $N$ sentences, let $\bar{e}=\frac{1}{N}\sum\limits_{s\in{D}}^{N}e_{s}$. Following~\cite{ChenFastAS2018,Wu2018LearningTE}, $d$ is then calculated as follows:
\begin{equation}\label{eq:doc rep}
  d=tanh\left(W_d\bar{e}+b_d\right),
\end{equation}

where $W_d\in\mathbb{R}^{n\times{n}}$ and $b_d\in\mathbb{R}^{n}$ are learnable parameters.

The second auxiliary representation is that of the summary that was generated by the editor so far, denoted at step $i$ as $g_{i-1}\in\mathbb{R}^{n}$, with $g_{0}=\vec{0}$. Such a representation provides a \textit{local context} for decision making. Given the four representations as an input, the editor's decision for sentence $s_{i}\in{S}$ is implemented using two fully-connected layers, as follows:
\begin{equation}\label{eq:action likelihood}
  softmax\left(Vtanh\left(W_c[e_{s_i},a_{s_i},g_{i-1},d]+b_c\right)+b\right),
\end{equation}

where $[\cdot]$ denotes the vectors concatenation, $V\in\mathbb{R}^{3\times{m}}$, $W_c\in\mathbb{R}^{m\times{4n}}$, $b_c\in\mathbb{R}^{m}$ and $b\in\mathbb{R}^{3}$ are learnable parameters.

In each step $i$, therefore, the editor chooses the action $\pi_i\in\{\textsf{E},\textsf{A},\textsf{R}\}$ with the highest likelihood (according to Eq.~\ref{eq:action likelihood}), further denoted $p(\pi_i)$. Upon decision, in case it is either \textsf{E} or \textsf{A}, the editor appends the corresponding sentence version (i.e., either $s_{i}^{e}$ or $s_{i}^{a}$) to $S'$; otherwise, the decision is \textsf{R} and sentence $s_{i}$ is discarded. Depending on its decision, the current summary representation is further updated as follows:
\begin{equation}\label{eq:doc rep}
  g_{i}=g_{i-1}+tanh\left(W_gh_i\right),
\end{equation}

where $W_g\in\mathbb{R}^{n\times{n}}$ are learnable parameters, $g_{i-1}$ is the summary representation from the previous decision step; and $h_i\in\{e_{s_i},a_{s_i},\vec{0}\}$, depending on which decision is made.

Such a network architecture allows to capture various complex interactions between the different inputs. 
For example, the network may learn that given the global context, one of the sentence versions may allow to produce a summary with a better coverage. As another example, based on the interaction between both sentence versions with either of the local or global contexts (and possibly among the last two), the network may learn that both sentence versions may only add superfluous or redundant information to the summary, and therefore, decide to reject both.

\subsection{Extractor and Abstractor}\label{sec:EA sum}
As a proof of concept, in this work, we utilize the extractor and abstractor that were previously used in~\cite{ChenFastAS2018}, with a slight modification to the latter, motivated by its specific usage within our approach. 
We now only highlight important aspects of these two sub-components and kindly refer the reader to~\cite{ChenFastAS2018} for the full implementation details.

The extractor of~\cite{ChenFastAS2018} consists of two main sub-components. The first is an \textit{encoder} which encodes each sentence $s\in{D}$ into $e_{s}$ using an hierarchical representation\footnote{\small Such a representation is basically a combination of a temporal convolutional model followed by a biLSTM encoder.}. The second is a \textit{sentence selector} using a \textit{Pointer-Network}~\cite{vinyals2015pointer}. For the latter, let $P(s)$ be the selection likelihood of sentence $s$.

The abstractor of~\cite{ChenFastAS2018} is basically a standard encoder-aligner-decoder with a copy mechanism~\cite{See2017GetTT}. Yet, instead of applying it directly only on a single given extracted sentence $s_{i}^{e}\in{S}$, we apply it on a ``chunk" of three consecutive sentences\footnote{\small The first and last chunks would only have two consecutive sentences.}  $(s_{-}^{e},s_{i}^{e},s_{+}^{e})$, where $s_{-}^{e}$ and $s_{+}^{e}$ denote the sentence that precedes and succeeds $s_{i}^{e}$ in $D$, respectively. This in turn, allows to generate an abstractive version of $s_{i}^{e}$ (i.e., $s_{i}^{a}$) that benefits from a wider local context. Inspired by previous soft-attention methods, we further utilize the extractor's sentence selection likelihoods $P(\cdot)$ 
for enhancing the abstractor's attention mechanism, as follows. Let $C(w_{j})$ denote the abstractor's original attention value of a given word $w_{j}$ occurring in $(s_{-}^{e},s_{i}^{e},s_{+}^{e})$; we then recalculate this value to be $C'(w_{j})=\frac{C(w_{j})\cdot P(s)}{Z}$, with $w_{j}\in{s}$ and $s\in\{s_{-}^{e},s_{i}^{e},s_{+}^{e}\}$; $Z=\sum_{s'\in\{s_{-}^{e},s_{i}^{e},s_{+}^{e}\}}\sum_{w_{j}\in{s'}}C(w_{j})\cdot P(s')$ denotes the normalization term.

\subsection{Sentence representation}
Recall that, in order to compare $s_{i}^{e}$ with $s_{i}^{a}$, we need to represent both sentence versions on as similar grounds as possible.
To achieve that, we first replace $s_{i}^{e}$ with $s_{i}^{a}$ within the original document $D$. By doing so, we basically treat sentence $s_{i}^{a}$ as if it was an ordinary sentence within $D$, where the rest of the document remains untouched. We then obtain $s_{i}^{a}$'s representation by encoding it using the extractor's encoder in a similar way in which sentence $s_{i}^{e}$ was originally supposed to be encoded.
This results in a representation $a_{s_i}$ that provides a comparable alternative to $e_{s_i}$, whose encoding is expected to be effected by similar contextual grounds.

\subsection{Network training}\label{sec:network_train}
We conclude this section with the description of how we train the editor using a novel soft labeling approach.
Given text $S$ (with $l$ extracted sentences), let $\pi=(\pi_{1},\ldots,\pi_{l})$ denote its editing decisions sequence. We define the following ``soft" cross-entropy loss:
\begin{equation}\label{eq:loss function}
  \mathcal{L}(\pi|S)=-\frac{1}{l}\sum\limits_{s_{i}\in{S}}\sum_{\pi_i\in\{\textsf{E},\textsf{A},\textsf{R}\}}y(\pi_i)\log p(\pi_i),
\end{equation}

where, for a given sentence $s_{i}\in{S}$, $y(\pi_i)$ denotes its soft-label for decision.  

We next explain how each soft-label $y(\pi_i)$ is estimated. 
To this end, we utilize a given summary quality metric $r(S')$ which can be used to evaluate the quality of any given summary $S'$ 
(e.g., ROUGE~\cite{lin2004rouge}). Overall, for a given text input $S$ with $l$ sentences, there are $3^{l}$ possible summaries $S'$ to consider. Let $\pi^{*}=(\pi_{1}^{*},\ldots,\pi_{l}^{*})$ denote the best decision sequence which results in the summary which maximizes $r(\cdot)$. For $i\in\{1,2,\ldots,l\}$, let $\bar{r}(\pi_{1}^{*},\ldots,\pi_{i-1}^{*},\pi_{i})$ denote the average $r(\cdot)$ value obtained by decision sequences that start with the prefix $(\pi_{1}^{*},\ldots,\pi_{i-1}^{*},\pi_{i})$. Based on $\pi^{*}$, the soft label $y(\pi_i)$ is then calculated\footnote{\small For $i=1$ we have: $\bar{r}(\pi_{1}^{*},\ldots,\pi_{0}^{*},\pi_{1})=\bar{r}(\pi_{1})$.} as follows:
\begin{equation}\label{eq:soft label}
  y(\pi_i)=\frac{\bar{r}(\pi_{1}^{*},\ldots,\pi_{i-1}^{*},\pi_i)}{\sum_{\pi_j\in\{\textsf{E},\textsf{A},\textsf{R}\}}\bar{r}(\pi_{1}^{*},\ldots,\pi_{i-1}^{*},\pi_j)}
\end{equation}

\section{Evaluation}
\subsection{Dataset and Setup}
We trained, validated and tested our approach using the non-annonymized version of the CNN/DailyMail dataset~\cite{Hermann:2015:TMR:2969239.2969428}. Following~\cite{DBLP:conf/conll/NallapatiZSGX16}, we used the story highlights associated with each article as its ground truth summary. We further used the F-measure versions of ROUGE-1 (R-1), ROUGE-2 (R-2) and ROUGE-L (R-L) as our evaluation metrics~\cite{lin2004rouge}.

The extractor and abstractor were trained similarly to~\cite{ChenFastAS2018} (including the same hyperparameters).
The Editorial Network (hereinafter denoted \textit{EditNet}) was trained according to Section~\ref{sec:network_train}, using the ADAM optimizer with a learning rate of $10^{-4}$ and a batch size of $32$. To speedup the training time, we precalculated the soft labels (see Eq.~\ref{eq:soft label}).
Following~\cite{DBLP:conf/emnlp/DongSCHC18,Wu2018LearningTE}, we set the reward metric to be $r(\cdot) = \alpha R\mbox{-}1(\cdot) + \beta R\mbox{-}2(\cdot) + \gamma  R\mbox{-}L(\cdot)$; with $\alpha=0.4$, $\beta=1$ and $\gamma=0.5$, which were further suggested by~\cite{Wu2018LearningTE}.

\begin{table}[tb!]
\centering
\caption{Quality evaluation using ROUGE F-measure (ROUGE-1, ROUGE-2, ROUGE-L) on CNN/DailyMail non-annonymized dataset}
\begin{tabular}{|l|c|c|c|}
\hline
& R-1 & R-2 & R-L \\ \hline
\multicolumn{4}{|c|}{\textbf{Extractive}}                                                                   \\ \hline
\begin{tabular}[c]{@{}l@{}}Lead-3  \end{tabular}  & 40.00   & 17.50   & 36.20   \\
\begin{tabular}[c]{@{}l@{}}SummaRuNNer \scriptsize {\cite{Nallapati2017a}} \end{tabular}  & 39.60   & 16.20   & 35.30   \\
\begin{tabular}[c]{@{}l@{}}Refresh \scriptsize {\cite{DBLP:conf/naacl/NarayanCL18}} \end{tabular} & 40.00    & 18.20    & 36.60    \\
\begin{tabular}[c]{@{}l@{}}Rnes w/o coherence \scriptsize {\cite{DBLP:conf/aaai/WuH18}} \end{tabular}             & 41.25   & 18.87   & 37.75   \\
\begin{tabular}[c]{@{}l@{}}BanditSum \scriptsize {\cite{DBLP:conf/emnlp/DongSCHC18}} \end{tabular}             & 41.50   & 18.70   & 37.60   \\
\begin{tabular}[c]{@{}l@{}}Latent \scriptsize {\cite{ZhangLatent2018}}\end{tabular} & 41.05   & 18.77   & 37.54   \\
\begin{tabular}[c]{@{}l@{}}rnn-ext+RL \scriptsize {\cite{ChenFastAS2018}} \end{tabular}  & 41.47   & 18.72   & 37.76   \\
\begin{tabular}[c]{@{}l@{}}NeuSum \scriptsize {\cite{Qingyu2018}} \end{tabular}  & 41.59   & 19.01   & 37.98   \\\hline
\multicolumn{4}{|c|}{\textbf{Abstractive}}                                                                  \\ \hline
\begin{tabular}[c]{@{}l@{}}Pointer-Generator \scriptsize {\cite{See2017GetTT}} \end{tabular}   & 39.53   & 17.28   & 36.38   \\
\begin{tabular}[c]{@{}l@{}}KIGN+Prediction-guide \scriptsize {\cite{Li2018KIGN}} \end{tabular}   & 38.95   & 17.12   & 35.68   \\
\begin{tabular}[c]{@{}l@{}}Multi-Task(EG+QG) \scriptsize {\cite{GuoSoftLayer2018}} \end{tabular}   & 39.81   & 17.64   & 36.54   \\
\begin{tabular}[c]{@{}l@{}}RL+pg+cbdec \scriptsize {\cite{JiangClosedBook2018}} \end{tabular}   & 40.66   & 17.87   & 37.06   \\
\begin{tabular}[c]{@{}l@{}}Saliency+Entail. \scriptsize {\cite{Pasunuru2018a}}\end{tabular}       & 40.43   & 18.00   & 37.10   \\
\begin{tabular}[c]{@{}l@{}}Inconsistency loss \scriptsize {\cite{HsuAUM2018}}\end{tabular}  & 40.68   & 17.97   & 37.13   \\
\begin{tabular}[c]{@{}l@{}}Bottom-up \scriptsize {\cite{Gehrmann2018BottomUpAS}} \end{tabular}            & 41.22   & 18.68   & 38.34   \\
\begin{tabular}[c]{@{}l@{}}rnn-ext+abs+RL \scriptsize {\cite{ChenFastAS2018}} \end{tabular}  & 40.04   & 17.61   & 37.59   \\\hline
\multicolumn{4}{|c|}{\textbf{Mixed Extractive-Abstractive}}                                                                               \\ \hline
\textbf{EditNet}                                                            & 41.42   & 19.03   & 38.36   \\ \hline
\end{tabular}
\label{tab:ResultROUGE-F}
\end{table}

We further applied the \textit{Teacher-Forcing} approach~\cite{lamb2016professor} during training, where we considered the true-label instead of the editor's decision (including when updating $g_{i}$ at each step $i$ according to Eq.~\ref{eq:doc rep}).
Following~\cite{ChenFastAS2018}, we set $m=512$ and $n=512$.
We trained for $20$ epochs, which has taken about $72$ hours on a single GPU. We chose the best model over the validation set for testing. Finally, all components were implemented in Python $3.6$ using the pytorch $0.4.1$ package.

\subsection{Results}
Table~\ref{tab:ResultROUGE-F} compares the quality of \textit{EditNet} with that of several state-of-the-art extractive-only or abstractive-only baselines. This includes the extractor (\textit{rnn-ext-RL}) and abstractor (\textit{rnn-ext-abs-RL}) components of~\cite{ChenFastAS2018} that we further utilized for implementing\footnote{\small The \textit{rnn-ext-RL} extractor results reported in Table~\ref{tab:ResultROUGE-F} are the ones that were reported by~\cite{ChenFastAS2018}. Training the public extractor released by these authors, we obtained the following significantly lower results: R-1:38.43, R-2:18.07 and R-L:35.37.} \textit{EditNet} (see again Section~\ref{sec:EA sum}).

Overall, \textit{EditNet} provides a highly competitive summary quality, where it outperforms all baselines in the R-2 and R-L metrics. On R-1, \textit{EditNet} outperforms all abstractive baselines and almost all extractive ones. 
Interestingly, \textit{EditNet}'s summarization quality is quite similar to that of \textit{NeuSum}~\cite{Qingyu2018}. Yet, while \textit{NeuSum} applies an extraction-only approach, summaries generated by \textit{EditNet} include a mixture of sentences that have been either extracted or abstracted.

On average, $56\%$ and $18\%$ of \textit{EditNet}'s decisions were to abstract (\textsf{A}) or reject (\textsf{R}), respectively. Moreover, on average, per summary, \textit{EditNet} keeps only 33\% of the original (extracted) sentences, while the rest (67\%) are abstracted ones. 
This demonstrates that, \textit{EditNet} has a high capability of utilizing abstraction, while being also able to maintain or reject the original extracted text whenever it is estimated to provide the best benefit for the summary's quality.

\section{Conclusion and Future Work}
We have shown that instead of solely applying extraction or abstraction, a better choice would be a mixed one. As future work, we plan to evaluate other alternative extractor+abstractor configurations and try to train the network end-to-end. We further plan to explore reinforcement learning (RL) as an alternative decision making approach.

\balance
\bibliographystyle{plain}

\end{document}